\def\year{2020}\relax
\documentclass[letterpaper]{article} 
\usepackage{aaai20}  
\usepackage{times}  
\usepackage{helvet} 
\usepackage{courier}  
\usepackage[hyphens]{url}  
\usepackage{graphicx} 
\usepackage{latexsym}
\usepackage{microtype}
\usepackage{url}
\usepackage{color,soul}
\usepackage{multirow}
\usepackage{graphicx}
\usepackage{booktabs}
\usepackage{amsmath}
\usepackage{xcolor}
\usepackage{tabu}
\usepackage{hyperref}
\usepackage[utf8]{inputenc}

\usepackage{graphicx}  
\newcommand{\kk}[1]{\textcolor{cyan}{\bf\small [#1 --KK]}}
\newcommand{\ab}[1]{\textcolor{blue}{\bf\small [#1 --AB]}}

\newcommand{\citet}[1]{\citeauthor{#1}~\shortcite{#1}}
\newcommand{\citep}{\cite}

\title{Towards Robust Toxic Content Classification}

\author{
Keita Kurita \\
  Carnegie Mellon University \\ Pittsburgh, PA 15213, USA \\
  {\tt kkurita@andrew.cmu.edu} \\ \And
  Anna Belova \\ 
  Carnegie Mellon University \\ Pittsburgh, PA 15213, USA \\
  {\tt abelova@alumni.cmu.edu} \\ \And
  Antonios Anastasopoulos \\ 
  Carnegie Mellon University \\ Pittsburgh, PA 15213, USA \\
  {\tt aanastas@andrew.cmu.edu} \\ 
 }

\date{}

\begin{document}
\maketitle
\begin{abstract}
  Toxic content detection aims to identify content that can offend or harm its recipients. Automated classifiers of toxic content need to be robust against adversaries who deliberately try to bypass filters. We propose a method of generating realistic model-agnostic attacks using a lexicon of toxic tokens, which attempts to mislead toxicity classifiers by diluting the toxicity signal either by obfuscating toxic tokens through character-level perturbations, or by injecting non-toxic distractor tokens.
  We show that these realistic attacks reduce the detection recall of state-of-the-art neural toxicity detectors, including those using ELMo and BERT, by more than~$50\%$ in some cases.
  We explore two approaches for defending against such attacks.
  First, we examine the effect of training on synthetically noised data.
  Second, we propose the Contextual Denoising Autoencoder (CDAE): a method for learning robust representations that uses character-level and contextual information to denoise perturbed tokens.\footnote{Our code is publicly available at \href{https://github.com/keitakurita/robust_toxicity_detection}{github.com/keitakurita/robust\_toxicity\_detection}.}
  We show that the two approaches are complementary, improving robustness to both character-level perturbations and distractors, recovering a considerable portion of the lost accuracy.
  Finally, we analyze the robustness characteristics of the most competitive methods and outline  practical considerations for improving toxicity detectors.
\end{abstract}


\section{Introduction}
Toxic content on the internet prevents the constructive exchange of ideas, excludes sensitive individuals from online dialogue, and inflicts mental and physical health impacts on the recipients. Notable examples of toxic content include hate speech and profanity. Given the sheer scale of internet communications, manual filtering of such content is difficult, requiring methods of automated filtering. 

\begin{figure}[t]
\centering
\includegraphics[scale=0.3]{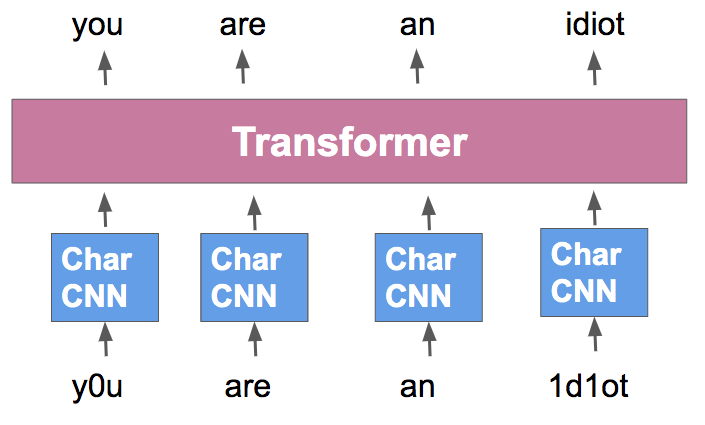}
\caption{Our proposed CDAE Architecture enhances the Transformer model with character level information, allowing for better handling of adversarial text.
}
\label{fig:cdae}
\end{figure}

Previous work in toxic content classification has so far focused on constructing classifiers that can flag toxic content with a high degree of accuracy on datasets curated from sources such as Twitter and Wikipedia. However, these datasets do not acknowledge the possibility for malicious users to attempt to deliberately bypass these classifiers.
In the presence of toxic content filters, these users could formulate \textit{adversarial attacks} that aim to prevent the classifier from detecting their harmful content while retaining readability for the receiving user. For example, a change of a single character to an asterisk, which requires minimal effort, may allow a hurtful content to bypass the toxic content filter (e.g., ``shut up" to ``s*ut up"). 

If such simple attacks are effective at fooling automated toxic content classifiers, the utility of these classifiers would diminish greatly: determined users could still easily produce toxic content at a large scale. 
Therefore, useful toxic content classifiers need to be robust to adversarial attacks by making the transmission of toxic content sufficiently difficult and discouraging users from posting this type of content.

In this paper, we investigate the robustness of state-of-the-art toxic content classifiers to realistic adversarial attacks as well as various defenses against them. We find that these classifiers are vulnerable to extremely simple, model-agnostic attacks, with the toxic comment recall rate dropping by nearly 50\% in some cases.

To address these vulnerabilities, we explore two types of defenses. The first is adversarial training, which we find to be effective against adversarial text, yet degrades performance on clean data. We also propose the Contextual Denoising Autoencoder (CDAE), a novel method for learning robust representations. The CDAE uses character-level \textit{and} contextual information to ``denoise'' obfuscated tokens. We find that our approach outperforms several strong baselines with respect to character-level obfuscations, but is still vulnerable to distractors (i.e., injected sequences of non-toxic tokens). We experimentally find that the two best-performing models (our proposed CDAE and BERT) have different robustness characteristics, but a model ensemble allows us to leverage both their advantages.

\section{Task and Datasets}\label{sec:data} 
Toxic content detection attempts to identify content that can offend or harm its recipients, including hate speech \cite{wang2018interpreting}, racism \cite{Waseem2016HatefulSO}, and offensive language \cite{wu2018decipherment}. Given the subjectivity of these categorizations, we do not limit the scope of our work to any specific type and address toxic content in general.
We work with three datasets summarized in Table~\ref{tab:stats}.

\begin{table}[t]
\small
\centering
\begin{tabular}{@{}c@{}cc@{}ccc@{}}
\toprule
\multirow{2}{*}{Dataset} & \multirow{2}{*}{Source} & \multirow{2}{*}{Usage} & \multicolumn{2}{c}{Examples} & \multirow{1}{*}{Toxic}  \\ 
 & & & Training & Test & (\%)\\
\midrule
{\small Jigsaw} & Wikipedia & Model & \multirow{2}{*}{159K} & \multirow{2}{*}{64K} & \multirow{2}{*}{9.5} \\
{\small 2018} & comments & train/eval &\\[.1em]
{\small Jigsaw} & Wikipedia & \multirow{2}{*}{Background} & \multirow{2}{*}{1.78M} & \multirow{2}{*}{n/a} & \multirow{2}{*}{5.9} \\
{\small 2019} & comments \\
{\small OffensEval} & \multirow{2}{*}{Twitter} & Model & \multirow{2}{*}{13K} & \multirow{2}{*}{860} & \multirow{2}{*}{33} \\
{\small 2019} & & train/eval &\\[.1em]
\bottomrule
\end{tabular}
\caption{Dataset Statistics}
\label{tab:stats} 
\end{table}

The Jigsaw 2018 dataset focuses on the general toxic content detection task and it is comprised of approximately 215,000 annotated comments from Wikipedia talk pages labeled by crowd workers. It provides both a general toxicity label and more fine-grained annotations such as severe toxicity, obscenity, threat, insult, and identity hate.

The Jigsaw Unintended Bias in Toxicity Classification dataset (Jigsaw 2019)\footnote{Publicly available at \href{https://www.kaggle.com/c/jigsaw-unintended-bias-in-toxicity-classification/overview}{https://www.kaggle.com/c/jigsaw-unintended-bias-in-toxicity-classification/overview}.} 
extends the Jigsaw 2018 dataset with 1.8 million comments, each annotated by up to 10 annotators for multiple labels.
Jigsaw 2019 contains a field for toxicity which provides the fraction of annotators who labeled the comment as ``toxic'' (7.99\% $\ge \!\! 0.5$). 
We use the Jigsaw 2019 corpus as our background corpus for generating adversarial attacks.

The OffensEval 2019\footnote{\href{https://competitions.codalab.org/competitions/20011}{https://competitions.codalab.org/competitions/20011}} dataset consists of 13,240 tweets annotated by crowdworkers. 
The data contains labels for whether the content is offensive and whether it is targeted, with 33\% of the tweets being labeled as offensive. 


\section{Generating Realistic Adversarial Attacks} \label{sec:noise}
Previous work generating adversarial examples in text often assumes access to either the weights \cite{Ebrahimi2017HotFlipWA,LiangLSBLS17} or the raw prediction scores of the classifier \cite{DBLP:journals/corr/abs-1812-05271,Liang:2018:DTC:3304222.3304355,alzantot2018,abs-1801-04354,SamantaM17}. However, it is unlikely that users would have access to this information. Instead, the users most likely would only have weak signals from what gets flagged as well as access to public datasets with toxicity labels. To mimic this setup, we use a large background corpus (Jigsaw 2019) with labels indicating toxicity.\footnote{We follow Jigsaw 2019 guidelines for conversion of their continuous toxicity scores into binary labels.} Our adversarial attack consists of two steps: (1) constructing a lexicon of toxic tokens and (2) using it to applying noise to the test set. 

\begin{figure}[t]
\includegraphics[scale=0.27]{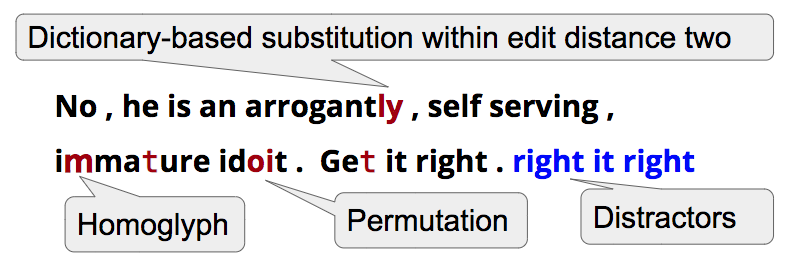}
\caption{An Example of a Noised Sentence \label{fig:noise}}
\end{figure}

To identify ``toxic'' tokens, we train a logistic regression classifier on bag-of-words utterance representations from our background corpus. We use the coefficients of the logistic regression classifier as a signed measure of the association between the token and toxicity and select the 50,000 tokens with the strongest positive association with toxicity to be our \textit{toxic lexicon}. 
We provide a list of top 100 toxic lexicon tokens in the Supplemental Material.
We treated any token that did not appear in our lexicon as non-toxic. Using this toxic  lexicon, we generate noised versions of the corpora using two settings: token obfuscation and distractor injection. Figure~\ref{fig:noise} provides an illustration of all our proposed attacks.

\subsection{Token Obfuscation} 
We apply character-level perturbations to 
the tokens of the utterance that belong
to our toxic lexicon. 
For each toxic token we randomly select one of the following three perturbing operations: character scrambling, homoglyph substitution, and dictionary-based near-neighbor replacement.
Details of the perturbing operations are given below.

\noindent \textbf{Character scrambling} consists in randomly permuting the token's characters 
without deletions and substitutions, as applied in other work \cite{heigold-etal-2018-robust,belinkov2018synthetic,michel2019}. Prior research shows that humans can read sufficiently long scrambled words, albeit not without an effort, especially if starting and ending letters remain in place \cite{rayner2006raeding}. Thus, for this operation, we ignore tokens with fewer than three characters and keep the first and the last character unchanged. The remaining characters are split into groups of three consecutive characters and each group is permuted randomly and independently.

\noindent \textbf{Homoglyph substitution} consists in replacing
one or more Latin letters with similar-looking international
characters from a handcrafted confusion map (see Supplemental Material). If homoglyph substitution operation is selected, each character of the toxic token is replaced with 20\% probability. This type of obfuscation is common in social media \cite{Rojas-Galeano:2017:OOO:3079924.3032963} and cybercrime \cite{ginsberg2018rapid,elsayed2018large}.

\noindent \textbf{Dictionary-based near-neighbor replacement} uses a base vocabulary
to find the closest (but distinct) token in terms of Levenshtein distance. If \emph{relative} Levenshtein distance (i.e., Levenshtein distance divided by maximum word length) is greater than 0.75, we use this nearest neighbor as a replacement. We leave the original toxic token unchanged otherwise. This form of noise produces common misspellings. As such, it introduces deletions, insertions, and substitutions that are not overly artificial. This procedure is distinct from that used by \citet{belinkov2018synthetic}, who generate naturally occurring errors using corpora of error corrections.  

\subsection{Distractor Injection} 
In this setting, we inject distractor tokens by repeating randomly selected sequences of non-toxic tokens. We split the utterance into two parts at a random position and find the maximum-length sequence of non-toxic words that starts in each of the parts. Search localization introduces variety in the identified distractor sequences, which helps to avoid the appearance of easily detectable vandalism.
Once a suitable sequence is found, it is appended to the end of the utterance. 

Both, token obfuscation and distractor injection are model-agnostic, simple, and subject to easy automation.
Hence, toxic content classifiers that are vulnerable to these attacks can be easily and systematically exploited. 

We emphasize that the noise we present here is \textbf{different} from ``naturally'' occurring noise (e.g., misspellings and slang) that does not deliberately attempt to hide toxic tokens. The datasets we use have not been constructed in the presence of a toxicity filter, implying that the users had no incentive to obfuscate toxicity of their comments. Hence, the synthetic noise we present here is not the noise that we observe frequently in these datasets. 


\section{Effect of Adversarial Noise}
We have implemented an experiment to assess whether our perturbations retained the toxicity of the toxic comments to human readers. In that, we have randomly sampled 200 comments from the Jigsaw 2018 dataset, half of which labeled as toxic by original Jigsaw 2018 crowd-workers. For each comment, we have a native English speaker rate either the perturbed or unperturbed version of the comment, taking care not to show both versions to the same individual. Overall, our experiment involved 10 participants, with each individual providing a toxicity rating for 80 comments. As such, we have obtained a total of 800 ratings, with each version of the comment receiving two independent ratings. We have tested whether the toxicity rating of the unperturbed comment tended to be higher than that of the perturbed comment using the Wilcoxon signed rank test \cite{wilcoxon1992individual} applied to pairs of unperturbed/perturbed toxicity scores averaged at the comment-level. The original comment was perceived as more toxic (based on the average rating of two distinct users) 14\% of the time and we found no statistically significant difference at the 1\% significance level. Thus, we conclude that it is unlikely that our perturbations remove the toxicity signals for human readers.

We evaluate the effect of our adversarial noise on toxic content classifiers on the Jigsaw 2018 and OffensEval 2019 datasets \footnote{Note that the Jigsaw 2018 and Jigsaw 2019 datasets are distinct and we remove all examples in the Jigsaw 2019 dataset from the Jigsaw 2018 dataset to prevent leakage. We use the Jigsaw 2019 dataset as a background corpus and not to train the model for the Jigsaw 2018 dataset.}
The general toxic content classifier architecture is straightforward. The tokens $x_1,\ldots, x_T$ of an utterance $\mathbf{X}$ are first embedded into a continuous vector space and then passed through an LSTM encoder which produces a sequence of intermediate representations $\mathbf{H}=h_1,\ldots,h_T$.
These representations are then used to produce a single vector representation $h_c$ using mean- and max-pooling as well as attention:
\[
h_c = [\textrm{maxpool}(\mathbf{H}); \textrm{meanpool}(\mathbf{H}); \textrm{attention}(\mathbf{H})]
\]
which is, in turn, put through an MLP and used to make a  prediction $\hat{y}$ of the toxicity of the utterance through a sigmoid function:
\[
\hat{y} = \sigma(\textrm{MLP}(h_c)).
\]

To demonstrate the effect of our adversarial attacks, we experiment with fastText \cite{bojanowski-etal-2017-enriching} and ELMo \cite{Peters:2018} embeddings, both of which are capable of handling out-of-vocabulary words. 
For ELMo, we follow the recommendations of \citet{Peters:2018} and apply a 0.5 dropout to the representations and a weight decay of 1e-4 to the scalar weights of all layers. We only fine-tune the scalar weights and keep the language model weights fixed.
We also experiment with BERT, applying a single affine layer to the embedding of the [CLS] token for classification and fine-tune all weights. In addition, we report the performance of a simple logistic regression baseline. 

All hyperparameters are tuned on the Jigsaw 2018 dataset and are listed in the Supplemental Material.
Preprocessing steps include tokenization, lower-casing, removal of hyperlinks and removal of characters that are repeated more than three times in a row (e.g., ``stupiiiiddddd'' is converted to ``stupid'', but ``iidiot'' remains unchanged). All punctuation is retained. For consistency across datasets, we evaluate models on the ``toxic''/``offensive'' labels that include all types of toxicity (obscenity, hate speech, targeted/untargeted offense, and others).
To convert probabilistic outputs of the models to binary classes, we threshold the predictions to maximize the F1 score on the training set.
We focus on the ability of various models to classify toxic content correctly since this is where adversarial attacks are most likely to take place (users that post non-toxic content are not motivated to have the system misclassify their content as toxic). 

\begin{table}[t]
\small
\centering
\begin{tabular}{l|rr|rr}
\toprule
\multicolumn{1}{l|}{\multirow{2}{*}{Noise}} & \multicolumn{2}{c}{Jigsaw 2018} & \multicolumn{2}{|c}{OffensEval 2019} \\ 
\multicolumn{1}{l|}{} & \multicolumn{1}{c}{Recall} & \multicolumn{1}{c}{\% Change} & \multicolumn{1}{|c}{Recall} & \multicolumn{1}{c}{\% Change} \\ \midrule
 & \multicolumn{4}{c}{Logistic Regression} \\ \midrule
None &  0.822 &  & 0.621 & \\
C+D & 0.344 & -58.2 &  0.246 & -60.4 \\
\midrule
 & \multicolumn{4}{c}{fastText-based model} \\ \midrule
None &  0.902 &  & 0.633 & \\
C+D & 0.485 & -46.2 &  0.350 & -44.7 \\
\midrule
 & \multicolumn{4}{c}{ELMo-based model} \\ \midrule
None &  0.887 &  & 0.596 & \\
C+D & 0.569 & -35.9 &  0.350 & -41.3 \\
\midrule
 & \multicolumn{4}{c}{BERT-based model} \\ \midrule
None &  0.914 &  & 0.721 & \\
C+D & 0.597 & -34.7 &  0.342 & -52.6 \\
 \bottomrule
\end{tabular}
\caption{\label{tab:noised_results} 
The combination of character-level perturbations (C) and distractors (D) leads to substantial loss of recall across all models on both test sets. 
}
\end{table}

The effects of our combined adversarial attacks are summarized in Table~\ref{tab:noised_results}. The logistic regression classifier is effectively incapable of handling out-of-vocabulary words and performs the worst when noise is applied, with more than~50\% recall lost. Despite this limitation, however, its performance does not drop to zero. This means that our obfuscation does not completely remove \textit{all} words that the logistic regression classifier uses to detect toxicity. Indeed, we found that some tokens that are quite obviously toxic (e.g., ``motherf*cker'') were not included in our toxic lexicon. Therefore, it is likely that improving the lexicon by finding a larger dataset or manually curating more toxic words could further enhance the effect of adversarial noise. Although neural models fare slightly better, recall on the adversarial test sets still drops significantly, with losses of over~30\% in all cases.

We present randomly sampled examples of toxic sentences that were misclassified by the fastText model due to the adversarial noise in Table~\ref{tab:examples}. Although not all of them retain grammatical correctness, it is our view that their toxicity is preserved and they should be properly handled by any toxic content classifier 

\begin{table*}[t]
\small
\centering
\begin{tabular}{ll}
\toprule
Original & Perturbed  \\ 
\midrule
\includegraphics[width=0.5\textwidth]{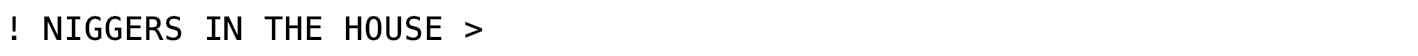} & \includegraphics[width=0.5\textwidth]{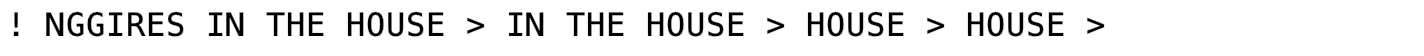} \\
\includegraphics[width=0.5\textwidth]{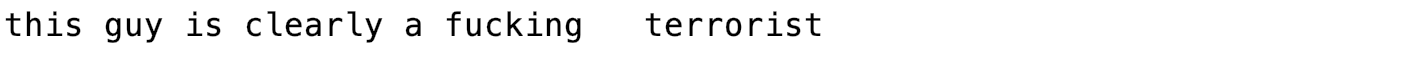} & \includegraphics[width=0.5\textwidth]{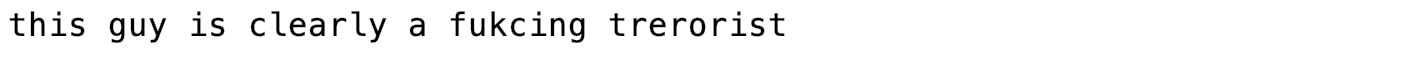} \\
\includegraphics[width=0.5\textwidth]{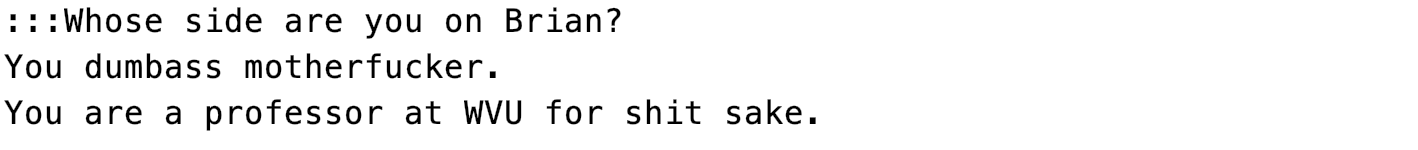} & \includegraphics[width=0.5\textwidth]{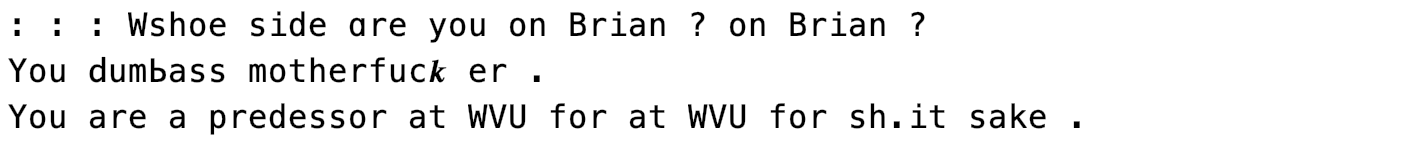} \\
\includegraphics[width=0.5\textwidth]{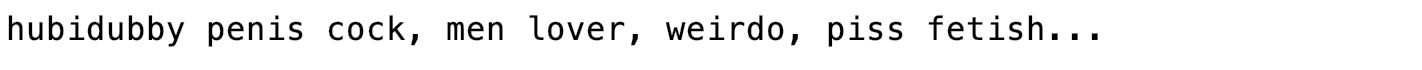} & \includegraphics[width=0.5\textwidth]{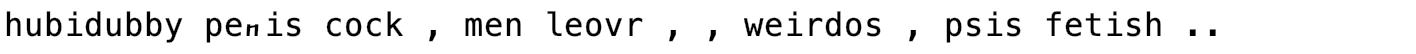} \\
\includegraphics[width=0.5\textwidth]{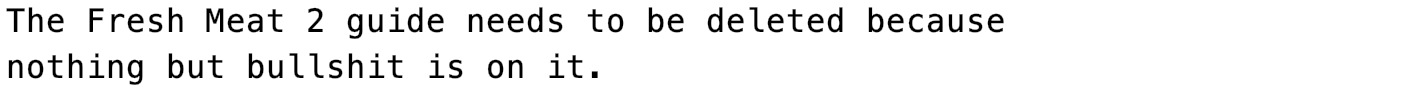} & \includegraphics[width=0.5\textwidth]{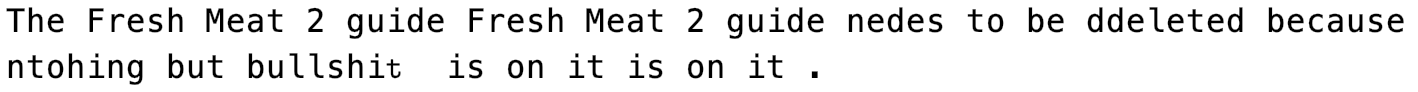} \\
\bottomrule
\end{tabular}
\caption{Examples of Toxic Sentences that were Misclassified by the fastText Model due to Adversarial Noise.}
\label{tab:examples} 
\end{table*}

\section{Defenses Against Adversarial Attacks}
Next we consider potential defenses against the aforementioned attacks: adversarial training and contextual denoising autoencoder. We note that our objective with these 

\subsection{Adversarial Training}
One possible defense is adversarial training \citep{Szegedy2014IntriguingPO,43405}, applying similar noise to the training dataset. Adversarial training has been applied successfully in tasks including machine translation \cite{belinkov2018synthetic} and morphological tagging \cite{heigold-etal-2018-robust}.
One limitation to this approach is that one would need to know the details of the incoming attack, \textbf{including the lexicon} the adversary might use to generate noise. This is a major limitation, since adversaries can easily change their lexicon. Another limitation is that there is no guarantee that the adversarial noise will produce a reliable pattern that the model can generalize to. 
For example, for fastText embeddings, the same operation of swapping two characters would produce completely different changes in the subwords for different source words, resulting in different changes in embedding space.
The model could also overfit to the adversarial noise, resulting in worsened performance on clean data.

\subsection{Contextual Denoising Autoencoder}
With token obfuscation, the underlying problem is that small character perturbations can cause large and unpredictable changes in embedding space. To resolve this problem, the underlying text representations themselves need to be robust against character-level perturbations. To learn such robust representations, we train a denoising autoencoder that receives noised tokens as input and predicts the denoised version of the token. When denoising tokens, the surrounding context can provide strong hints as to what the original token was. Some words like ``duck'' can be used both as obfuscations of profanity and as standard language, meaning context is crucial in effective denoising. Thus, we use a model that takes the context a sequence of potentially noised tokens as input and predicts the denoised tokens using contextual information. We call this model the \textit{Contextual Denoising Autoencoder} (CDAE). 

Due to its impressive performance across a wide range of tasks, we use a Transformer \cite{46201} as the underlying architecture. For word representations, we employ the character convolutional neural network (CNN) encoder used in the ELMo model. We feed the outputs of the CNN encoder to the Transformer with learned positional embeddings, 6 layers and 4 attention heads in each layer where the outputs of each layer are of size 128. We show the overall scheme of the CDAE in Figure \ref{fig:cdae}.
Not using wordpieces leads to massive vocabulary size, especially with corpora obtained from the web. We therefore use the CNN-softmax method combined with importance sampling loss \cite{45446} to accelerate training. We apply noise to 70\% of tokens according to the scheme in Section \ref{sec:noise} and mask all tokens uniformly with a probability of 10\%. We train our denoising autoencoder on a random subset of the UMBC webbase corpus \cite{UMBC} (a large-scale corpus constructed from web crawls) and the Jigsaw 2019 dataset, taking care to remove any examples from the Jigsaw 2018 dataset.

We note that this approach \textit{does} require knowledge of what character-level perturbations will be applied. However, the space of possible character-level perturbations that retain readability of the original token is limited. Crucially, unlike adversarial training, the CDAE does not require knowledge of the adversary's lexicon, making this approach more suitable for a wider range of attacks.

\begin{table*}[t]
\small
\begin{center}
\begin{tabular}{lll|ccc|ccc}
\toprule \multirow{2}{*}{Model} & \multirow{2}{*}{Train Noise} & \multirow{2}{*}{Test Noise} & \multicolumn{3}{|c}{Jigsaw 2018} & \multicolumn{3}{|c}{OffensEval 2019}\\
 & & & AUC  & F1  & Recall  & AUC  & F1  & Recall   \\ 
\midrule
\multirow{1}{*}[-.4em]{Logistic} & None & None & 0.959 & 0.652 & 0.822  & 0.813 & 0.619 & 0.621 \\ 
\multirow{2}{*}{Regression} & None & C+D & 0.877 & 0.459 & 0.344 & 0.639 &	0.340 &	0.246\\ 
                          & C+D & C+D & 0.906 & 0.565 & 0.632  & 0.682 & 0.432 & 0.408\\
                          \midrule
\multirow{3}{*}{FastText} & None & None & 0.973	 & 0.674 &	0.902  & 0.850 & 0.670 & 0.633 \\ 
                          & None & C+D & 0.905 & 0.546 & 0.485 & 0.755 & 0.450 & 0.350\\
                          & C+D & C+D & 0.932 & 0.591 & 0.643  & 0.763 & 0.532 & 0.540 \\
                          \midrule
\multirow{3}{*}{ELMo} & None & None & 0.970 & 0.654 & 0.887  & 0.843 & 0.644 & 0.596 \\
                      & None & C+D & 0.880 & 0.538 & 0.569 & 0.725 & 0.429 & 0.350  \\
                      & C+D & C+D & 0.917 & 0.568 & 0.696 & 0.759 & 0.549 & 0.579 \\    
                        \midrule
\multirow{3}{*}{BERT} & None & None & \textbf{0.974} & \textbf{0.685} & \textbf{0.914}  & \textbf{0.889} & \textbf{0.730} & 0.721\\
                      & None & C+D & 0.901 & 0.596 & \textbf{0.604} & 0.734 & 0.462 & 0.342 \\
                      & C+D & C+D & \textbf{0.940} & \textbf{0.614} & \textbf{0.765} & \textbf{0.769} & \textbf{0.554} & 0.446\\
                          \midrule
\multirow{3}{*}{CDAE (ours)} & None & None & 0.973 & 0.677 & 0.894  & 0.861 & 0.665 & \textbf{0.742}\\
                             & None & C+D & \textbf{0.918} & \textbf{0.597} & 0.597 & \textbf{0.747} & \textbf{0.479} & \textbf{0.388} \\
                             & C+D & C+D & 0.932 & 0.604 & 0.733 & \textbf{0.769} & 0.547 & \textbf{0.596} \\
\bottomrule
\end{tabular}
\end{center}
\caption{\label{tab:results} Detailed results of models on two datasets. The best results for each setting are \textbf{highlighted}. C+D refers to character-level perturbations and distractors combined. BERT generally performs strongest in clean settings. The CDAE is better at handling noise without adversarial training, while BERT and CDAE perform comparably when adversarial training is introduced. 
}
\end{table*}

\section{Effect of Defenses} \label{sec:results}

In order to evaluate our proposed defenses, we measure AUC, F1 score, and recall over the toxic class for all models.
The model architecture for CDAE is similar to the one we used for fastText and ELMo. For CDAE, we use the mean of the final 4 layers of the model and concatenate them with fastText embeddings, because we found that this leads to superior performance.\footnote{We hypothesize that this is because the fastText embeddings were trained on much more data so captured some semantic aspects that the CDAE did not.} The detailed results of applying adversarial training as well as CDAE's performance on the Jigsaw 2018 and OffensEval 2019 datasets are shown in Table~\ref{tab:results}.
\footnote{The OffensEval challenge evaluates models with a macro-averaged F1 score over both classes, so our numbers are significantly lower than the numbers reported there. We achieve a 0.84  macro-averaged F1 score, beating the state-of-the-art.}

Overall, we find that BERT performs well in the absence of noise on both datasets (None--None setting). As expected, the addition of noise hurts its performance. CDAE, on the other hand, performs well in the noised test set without adversarial training (None--C+D setting), indicating that it indeed manages to at least partly denoise the adversarial utterances. When additional adversarial training is introduced (C+D--C+D setting), BERT and CDAE perform comparably, outperforming all other methods. For OffenEval, we found that BERT was more biased towards the non-toxic class compared to the CDAE, causing it to have much higher precision but slightly lower recall.

Adversarial training improves performance across the board, although performance does not recover to the clean-data standards. Interestingly, classifiers that were more vulnerable to attacks before adversarial training tend to perform poorly even with adversarial training. This implies that the representation of text needs to be inherently robust for adversarial training to be an effective defense.
Despite using character CNNs, ELMo was more vulnerable to noise compared to our CDAE (cf. 36\% vs. 33\% degradation in recall on the Jigsaw 2018 dataset without adversarial training), showing that character CNNs need to be explicitly trained to handle noise/out-of-vocabulary words in order to exhibit robustness.

To better understand the robustness characteristics of our two best models (BERT and CDAE) models, we perform ablations under various noise settings (only character perturbations, only distractors, adversarial training with a clean test set). Results are shown in Table~\ref{tab:detresults_merge} and we summarize our findings below.

\begin{table*}[t]
\small
\begin{center}
\begin{tabular}{@{}lll|ccc|ccc@{}}
\toprule
\multirow{2}{*}{Model} &
\multicolumn{1}{l}{\multirow{2}{*}{Train Noise}} & \multicolumn{1}{l|}{\multirow{2}{*}{Test Noise}} & \multicolumn{3}{c}{Jigsaw 2018} & \multicolumn{3}{|c}{OffensEval 2019} \\ \cmidrule(l){4-9} 
& &  & \multicolumn{1}{c}{AUC} & \multicolumn{1}{c}{F1} & \multicolumn{1}{c}{Recall} & \multicolumn{1}{|c}{AUC} & \multicolumn{1}{c}{F1} & \multicolumn{1}{c}{Recall} \\ \midrule
\multirow{10}{*}{BERT} &
None & C & 0.911 & 0.600 & 0.608 & 0.757 & 0.446 & 0.313 \\
 & &  & -6.44\% & -12.3\% & -33.4\% & -14.8\% & -38.8\% & -56.6\% \\[1ex]
&C & C & 0.922 & 0.588 & 0.697 & 0.783 & 0.618 & 0.529 \\
& &  & -5.37\% & -14.0\% & -23.8\% & -11.9\% & -15.3\% & -26.5\% \\[1ex]
&None & D & \textbf{0.970} & \textbf{0.682} & \textbf{0.882} & \textbf{0.881} & \textbf{0.705} & 0.650 \\
& &  & -0.47\% & -0.41\% & -3.56\% & -0.97\% & -3.49\% & -9.83\% \\[1ex]
&D & D & \textbf{0.971} & \textbf{0.683} & \textbf{0.904} & \textbf{0.885} & \textbf{0.714} & 0.650 \\
& &  & -0.33\% & -0.20\% & -1.10\% & -0.50\% & -2.21\% & -9.83\% \\[1ex]
&C + D & None & \textbf{0.968} & \textbf{0.653} & 0.904 & \textbf{0.875} & \textbf{0.685} & \textbf{0.739} \\
& &  & -0.59\% & -4.60\% & -1.08\% & -1.64\% & -6.16\% & -11.4\% \\ \midrule 
\multirow{10}{*}{CDAE}
&None & C & \textbf{0.925} & \textbf{0.610} & \textbf{0.642} & \textbf{0.758} & \textbf{0.465} & \textbf{0.354} \\
& &  & -4.90\% & -9.87\% & -28.1\% & -11.9\% & -30.1\% & -52.2\% \\[1ex]
&C & C & \textbf{0.932} & \textbf{0.596} & \textbf{0.706} & \textbf{0.786} & \textbf{0.629} & \textbf{0.533} \\
& &  & -4.18\% & -11.9\% & -21.0\% & -8.72\% & -5.47\% & -28.0\% \\[1ex]
&None & D & 0.965 & 0.669 & 0.810 & 0.864 & 0.672 & \textbf{0.708} \\
& &  & -0.75\% & -1.20\% & -9.40\% & 0.36\% & 0.98\% & -4.50\% \\[1ex]
&D & D & 0.970 & 0.672 & 0.882 & 0.862 & 0.691 & \textbf{0.683} \\
& &  & -0.25\% & -0.74\% & -1.39\% & 0.15\% & 3.78\% & -7.87\% \\[1ex]
&C + D & None & \textbf{0.968} & 0.651 & \textbf{0.912} & 0.858 & 0.655 & 0.721 \\
& &  & -0.54\% & -3.84\% & 2.01\% & -0.35\% & -1.50\% & -2.81\% \\
 
 \bottomrule
\end{tabular}
\caption{\label{tab:detresults_merge} Detailed Results for the CDAE and BERT (Percentages below are changes relative to the unnoised baseline). C refers to character-level perturbations and D refers to distractors. Numbers in \textbf{bold} are best results for each setting/metric. Adversarial training helps most against character-level perturbations.  The CDAE is stronger against character-level perturbations, whereas BERT performs well in the presence of distractors. 
}
\end{center}
\end{table*}

\noindent \textbf{Character-level perturbation degrades performance more than distractors.}
For both datasets and models, character-level perturbations lead to significantly larger drops in performance across all metrics. This is reasonable, given that obfuscation directly removes the toxicity signal. The distractors, instead, simply dilute it.

\noindent \textbf{Adversarial training reduces performance on clean data.}
Although adversarial training consistently improves robustness to noise, it also slightly reduces performance on clean data. This undesirable byproduct can probably be attributed to models overfitting to the training noise. 

\noindent \textbf{The CDAE is more resilient against character perturbations compared to BERT.}
We find that the performance of the CDAE drops less with character-level perturbations both before and after adversarial training. For example the recall drops by 33\%  and 24\% for BERT before and after adversarial training, whereas for the CDAE the recall drop is 28\% and 21\% respectively. This reveals the advantage of the CDAE: it is explicitly trained to address character-level perturbations. BERT's vulnerability to such noise cannot be easily remedied due to its reliance to a wordpiece tokenizer.

\noindent \textbf{BERT performs better in the presence of distractors compared to the CDAE.}
In contrast to the CDAE, BERT is weak to character-level perturbations but strong against distractors. For both datasets, BERT performs stronger in terms of final performance, aside from recall on OffensEval where BERT was more inclined to predict the non-toxic class compared to the CDAE for all settings. For the Jigsaw dataset, BERT performance drops less in relative terms although the opposite holds for OffensEval. For OffensEval, the distractors tended to be shorter compared to the Jigsaw dataset since the original text was also generally shorter. This difference in response to distractors may suggest that BERT and the CDAE have different robustness characteristics regarding distractors. A possible explanation might lie in the architecture: BERT is entirely self-attention-based while the CDAE features are fed into a recurrent LSTM. The effect of the different architectures on the robustness characteristics towards distractors remains an open question.

\begin{table*}[t]
\small
\begin{center}
\begin{tabular}{lcc|ccc|ccc}
\toprule
\multicolumn{1}{l}{\multirow{2}{*}{Model}} &\multicolumn{1}{l}{\multirow{2}{*}{Train Noise}} & \multicolumn{1}{l|}{\multirow{2}{*}{Test Noise}} & \multicolumn{3}{c}{Jigsaw 2018} & \multicolumn{3}{|c}{OffensEval 2019} \\ \cmidrule(l){4-9} 
\multicolumn{1}{l}{}&\multicolumn{1}{l}{} & \multicolumn{1}{l|}{} & \multicolumn{1}{c}{AUC} & \multicolumn{1}{c}{F1} & \multicolumn{1}{c}{Recall} & \multicolumn{1}{|c}{AUC} & \multicolumn{1}{c}{F1} & \multicolumn{1}{c}{Recall} \\ \midrule

\multirow{2}{*}{BERT}
& None & C+D & 0.901 & 0.596 & 0.604 & 0.734 & 0.462 & 0.342 \\
& C+D & C+D & 0.940 & 0.614 & 0.765 & 0.769 & 0.554 & 0.446\\
  \midrule
\multirow{2}{*}{CDAE (ours)} 
& None & C+D & 0.918 & \textbf{0.597} & 0.597 & 0.747 & 0.479 & 0.388 \\
& C+D & C+D & 0.932 & 0.604 & 0.733 & 0.769 & 0.547 & 0.596 \\ 
\midrule
\multirow{2}{*}{Ensemble} 
& None & C + D & \textbf{0.921} & 0.590 & \textbf{0.725} & \textbf{0.774} & \textbf{0.505} & \textbf{0.404} \\
& C + D & C + D & \textbf{0.942} & \textbf{0.628} & \textbf{0.799} & \textbf{0.827} & \textbf{0.612} & \textbf{0.604} \\

\bottomrule
\end{tabular}
\end{center}
\caption{ \label{tab:ensemble} Results of the Ensemble of CDAE and BERT. The ensemble performs the strongest against combined noise among all of our methods. 
}
\end{table*}

\paragraph{Ensembling.} Based on our findings, we also examine the performance of an ensemble of BERT and the CDAE, in the hope that it will combine their advantages. The final prediction is made with arithmetic mean of the two models' predicted probabilities. Results are shown in Table~\ref{tab:ensemble}. 
Indeed, the ensemble outperforms both the single CDAE and BERT models when tested on combined noise, exactly because it combines their different robustness characteristics.
This suggests that although it may be difficult to train a single model to be robust to all possible attacks, specialized models can be trained to handle different attacks and their ensemble may be a simple, cheap approach that will boost robustness of the entire system.

\section{Related Work}

\paragraph{Toxic Content Classification.} Since toxic content classification is a text classification task, traditional techniques ranging from bag of words models \citep{georgakopoulos2018convolutional} to CNNs \citep{georgakopoulos2018convolutional} and RNNs \citep{van2018challenges,gunasekara2018review} have all been applied. 
Both \citet{van2018challenges} and \citet{gunasekara2018review} have shown that among the various approaches, bidirectional RNNs with attention using pretrained fastText embeddings \citep{joulin2016bag} have strong performance, with \citet{gunasekara2018review} acheiving the best single-model performance on the Jigsaw 2018 dataset using a bidirectional LSTM with attention. 
\citet{mishra-etal-2018-neural} developed an approach that uses a character-level models to mimic GloVe \citep{pennington-etal-2014-glove} embeddings, thus inferring the embeddings for unseen words. Crucially, this method can only train the model on in-vocabulary words, meaning it is incapable of handling targeted character obfuscations that do not appear naturally in the GloVe vocabulary.

\paragraph{Noise and Adversarial Attacks in Text.} 
\citet{belinkov2018synthetic} demonstrated the brittleness of neural machine translation (NMT) systems to both natural and synthetic noise. They showed that training on synthetically noised data improves robustness towards similar synthetic noise but not to naturally occurring noise (e.g. omissions). In contrast to their work, we focus on targeted adversarial attacks that deliberately attempt to fool a classifier. Multiple works have proposed white-box attacks (attacks assuming access to model gradients) in NLP for tasks such as NMT \cite{ebrahimi2018} and text classification \cite{Ebrahimi2017HotFlipWA}. \citet{SamantaM17} constructed a lexicon of words to construct adversarial examples, which is similar but crucially different from our approach in that they assume access to model gradients. Other work has explored black-box attacks \cite{Liang:2018:DTC:3304222.3304355}. In particular, \citet{HosseiniKZP17} generated adversarial attacks against the Google Perspective API, a public API for detecting toxic content, and showed the brittleness of this system. However, these methods rely on multiple queries to the underlying prediction scores of the model which are not always exposed to a user and can be seen as a form of internal knowledge. \citet{W16-5603} showed that the gender of posters on social media can be obfuscated by using a background corpus to identify words indicative of each gender and replacing those words with semantically similar words. Our method differs in that we replace words with similar-looking instead of similar-meaning character sequences, since our aim is to fool the system while maintaining readability. 

\noindent \textbf{Defenses.} One straightforward, yet non-scalable approach to solving the problem of adversarial noise is to manually curate a lexicon of the most frequent obfuscations \cite{Wang:2014:CET:2531602.2531734}.
On the other hand, \citet{Rojas-Galeano:2017:OOO:3079924.3032963} proposed a method to automatically match obfuscated words to their original forms using a custom edit distance function. Although their approach is more scalable, it still requires the manual construction of inflexible rules for measuring the distance caused by different transformation and thus can easily be circumvented by adversaries. \citet{W17-3005} proposed to use the errors from a class-conditioned character-level language model to classify out-of-vocabulary words as toxic or non-toxic. \citet{scrnn} proposed the semi-character level recurrent neural network (scRNN) as a method of generating robust word representations. Although their method showed strong performance in spell checking, it is unable to handle anagrams (e.g. ``there'' and ``three'') and homoglyph substitutions, and ignores contextual information.
One limitation of these approaches is that they do not consider the possibility of toxic words being mapped to in-vocabulary words. For instance, ``suck my duck'' is likely an obfuscation, but the word ``duck'' itself is common. These problems require the usage of context: for instance, ``duck'' in ``the duck is swimming'' is not toxic, but this can only be inferred based on the context. Moreover, none of these approaches consider distractor injection as a potential method of attack.


\section{Conclusion and Future Work}
In this paper, we show that we can easily degrade the performance of state-of-the-art toxic content classifiers in a model-agnostic manner by using a background corpus of toxicity and introducing character-level perturbations as well as distractors. We also explore defenses against these attacks, and we find that adversarial training improves robustness in general, but decreases performance on clean data. We also propose the Contextual Denoising Auto-Encoder (CDAE), a method of learning robust representations, and show that these representations are more robust against character-level perturbations, whereas a BERT-based model performs strongly in the presence of distractors. An ensemble of BERT and the CDAE is the most robust approach towards combined noise.



\bibliography{aaai_ref}
\bibliographystyle{aaai}

\appendix

\section*{Appendix}

\begin{table}[h]
\begin{center}
\caption{Homoglyph Confusion Map}\label{tab:Homogplyphs}
\includegraphics[scale=0.45]{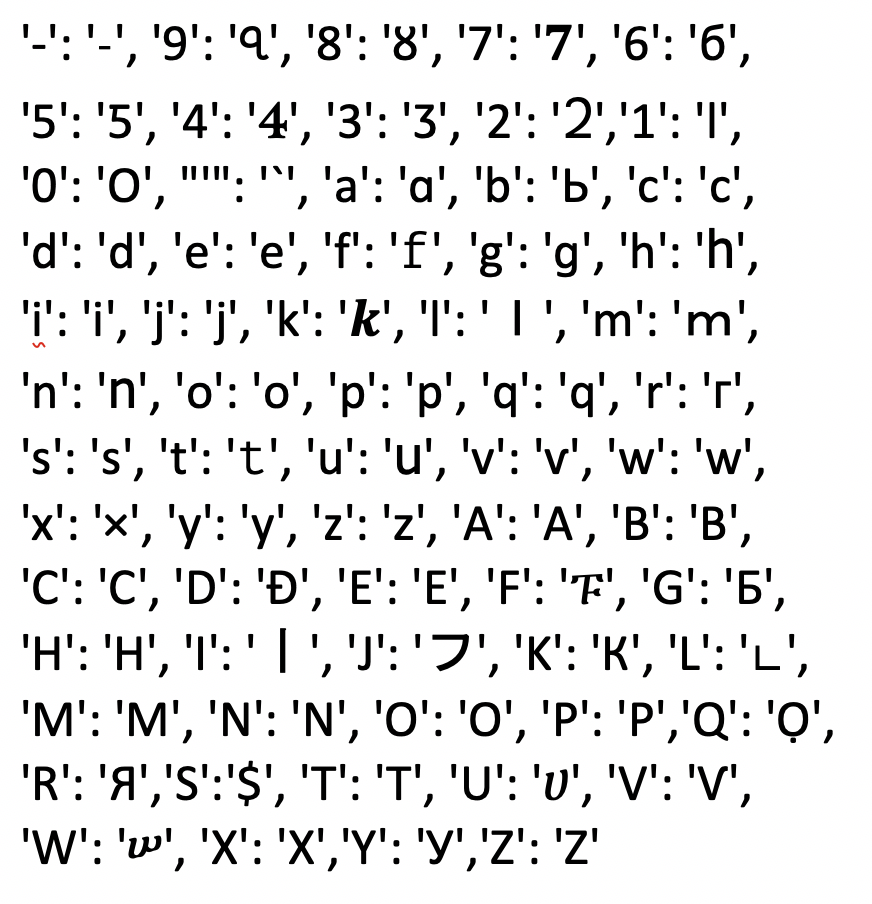}
\end{center}
\end{table}

\begin{table}[h]
\begin{center}
\caption{Hyperparameter Settings for BiLSTM model}\label{tab:HyperParams}
\begin{tabular}{ll}
\hline
Parameter & Baseline \\
\hline
Learning rate & 4e-3 \\
Batch size & 128 \\
Training epochs & 5 \\
Input dropout & 0.2\\
Word dropout & 0.1 \\
Output dropout & 0.3\\
Learning rate schedule & slanted\_triangular \\
\hline
\end{tabular}
\end{center}
\end{table}

\begin{table}[h]
\begin{center}
\caption{Hyperparameter Settings for BERT}\label{tab:HyperParams}
\begin{tabular}{ll}
\hline
Parameter & Baseline \\
\hline
Learning rate & 3e-5 \\
Batch size & 32 \\
Training epochs & 4 \\
Learning rate schedule & slanted\_triangular \\
\hline
\end{tabular}
\end{center}
\end{table}

\begin{table*}[h]
\small
\begin{center}
\caption{Top 100 Tokens of the Toxic Lexicon}\label{tab:ToxicLex}
\begin{tabular}{llllllll}
\hline
Rank & Token & Rank & Token & Rank & Token & Rank & Token \\
\hline
1  & idiot     & 26 & moronic    & 51 & whores       & 76  & stupider     \\
2  & idiots    & 27 & jerk       & 52 & dimwit       & 77  & pussies      \\
3  & stupidity & 28 & assholes   & 53 & f**k         & 78  & asses        \\
4  & stupid    & 29 & dumbass    & 54 & suck         & 79  & daft         \\
5  & idiotic   & 30 & scum       & 55 & suckers      & 80  & schmucks     \\
6  & shit      & 31 & jackass    & 56 & buffoon      & 81  & anal         \\
7  & asshole   & 32 & fools      & 57 & imbecilic    & 82  & retarded     \\
8  & morons    & 33 & jerks      & 58 & dumber       & 83  & asinine      \\
9  & stupidest & 34 & dumb       & 59 & cretin       & 84  & nitwit       \\
10 & imbecile  & 35 & damn       & 60 & stupidly     & 85  & parasites    \\
11 & bullshit  & 36 & hypocrites & 61 & loser        & 86  & p***y        \\
12 & moron     & 37 & bastards   & 62 & clowns       & 87  & fucked       \\
13 & imbeciles & 38 & bastard    & 63 & bitches      & 88  & wtf          \\
14 & fuck      & 39 & sucks      & 64 & shitty       & 89  & slut         \\
15 & bitch     & 40 & dammit     & 65 & ridiculous   & 90  & pigs         \\
16 & hypocrite & 41 & penis      & 66 & clown        & 91  & buffoons     \\
17 & idiocy    & 42 & ignorant   & 67 & silly        & 92  & testicles    \\
18 & scumbag   & 43 & arse       & 68 & coward       & 93  & dork         \\
19 & fucking   & 44 & foolish    & 69 & sucker       & 94  & troll        \\
20 & dumbest   & 45 & darn       & 70 & garbage      & 95  & disgusting   \\
21 & whore     & 46 & sh*t       & 71 & schmuck      & 96  & f*ck         \\
22 & crap      & 47 & rubbish    & 72 & damned       & 97  & liar         \\
23 & pussy     & 48 & scumbags   & 73 & hypocritical & 98  & dumbed       \\
24 & ass       & 49 & vagina     & 74 & lunatics     & 99  & cursed       \\
25 & pathetic  & 50 & fool       & 75 & losers       & 100 & masturbation \\
\hline
\end{tabular}
\end{center}
\end{table*}

\begin{table*}
\centering
\begin{tabular}{cc}
    \toprule
    Original & Perturbed \\ \midrule
    \multicolumn{2}{c}{ \includegraphics[width=.8\textwidth]{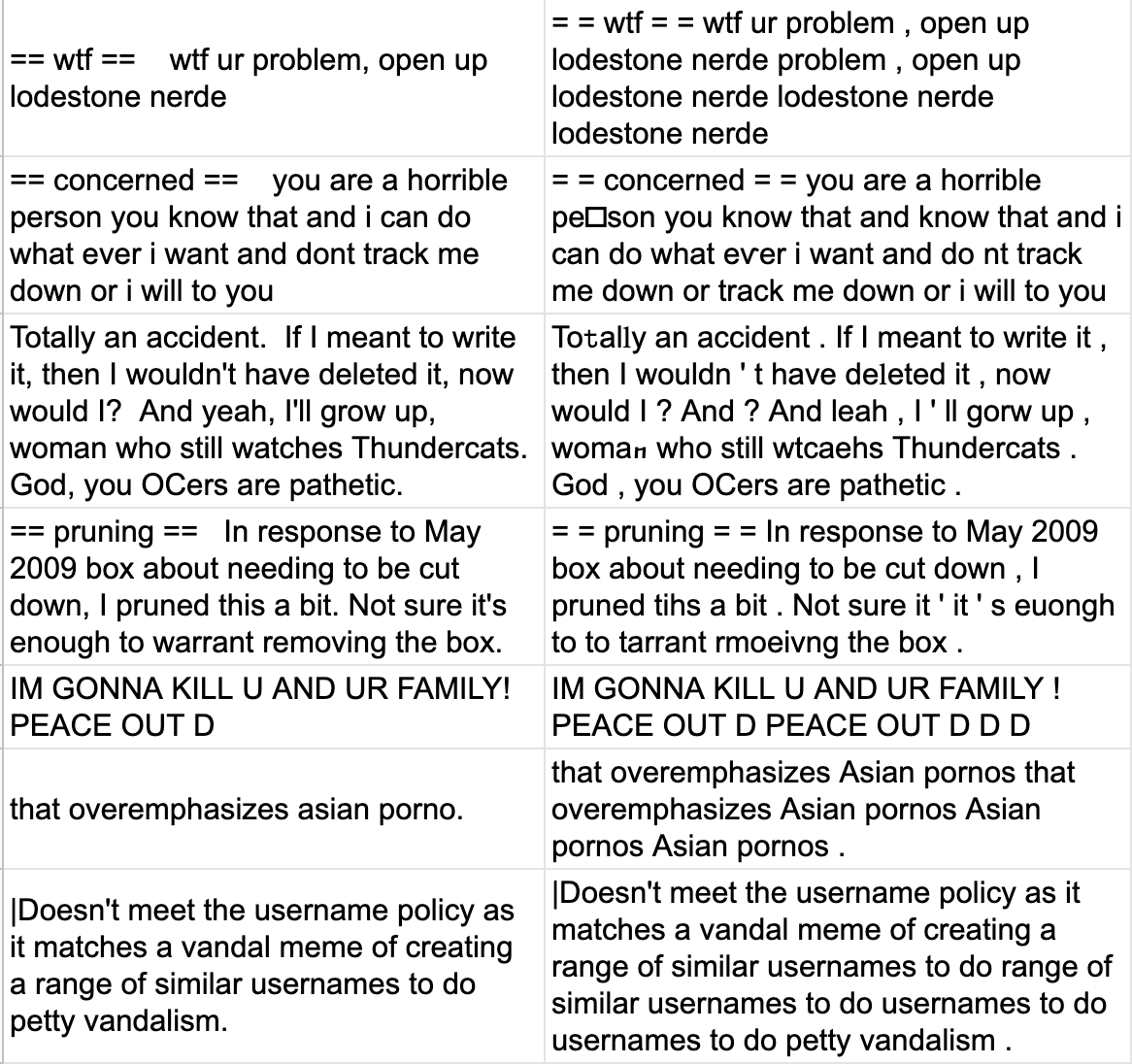} } \\
    \bottomrule
\end{tabular}
\label{app:ex1}
\caption{Examples of original and perturbed comments (part 1).}
\end{table*}
\begin{table*}
\centering
\begin{tabular}{cc}
    \toprule
    Original & Perturbed \\ \midrule
    \multicolumn{2}{c}{ \includegraphics[width=.8\textwidth]{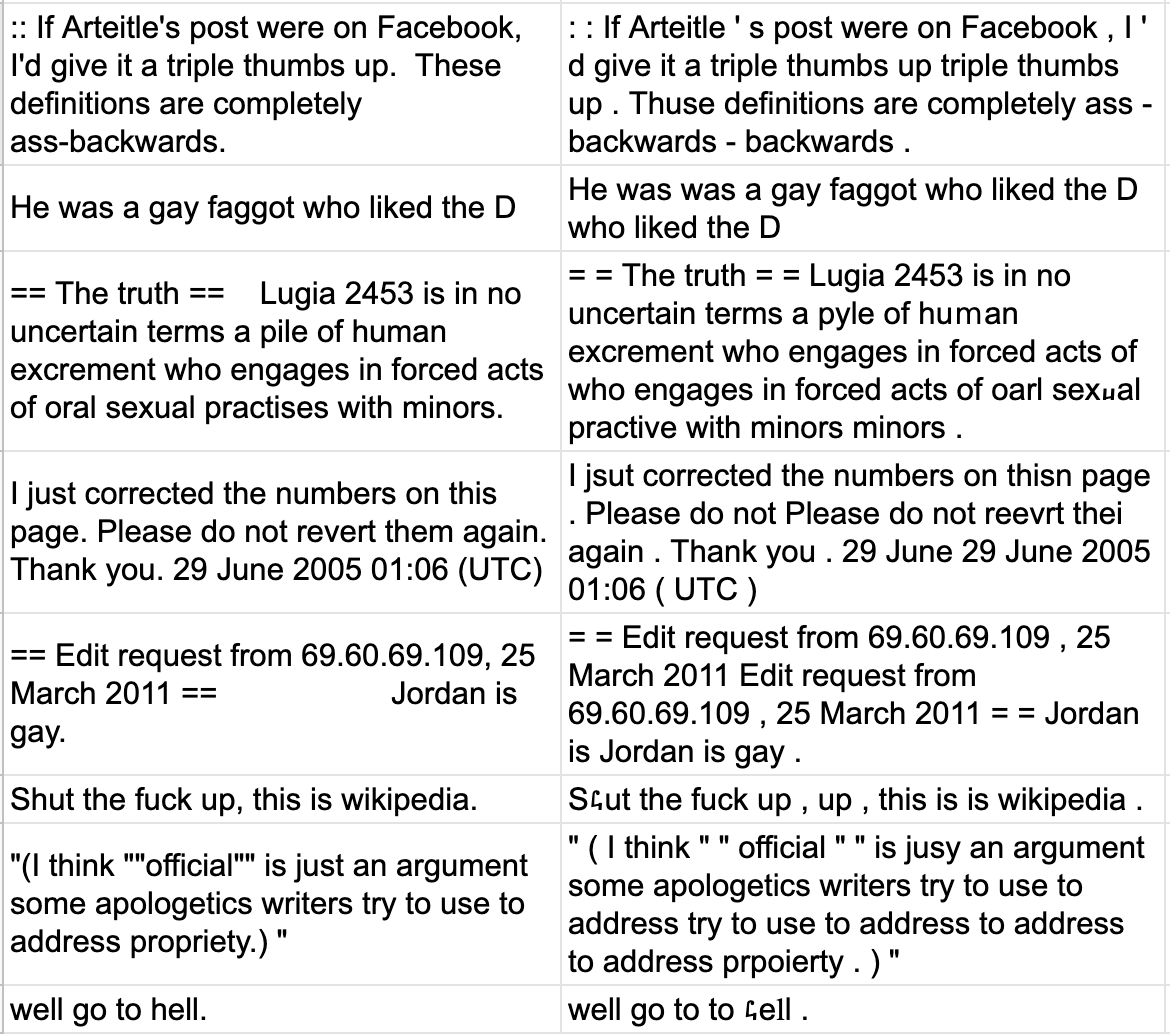} } \\
    \bottomrule
\end{tabular}
\label{app:ex2}
\caption{Examples of original and perturbed comments (part 2).}
\end{table*}

\begin{table*}
\centering
\begin{tabular}{cc}
    \toprule
    Original & Perturbed \\ \midrule
    \multicolumn{2}{c}{ \includegraphics[width=.8\textwidth]{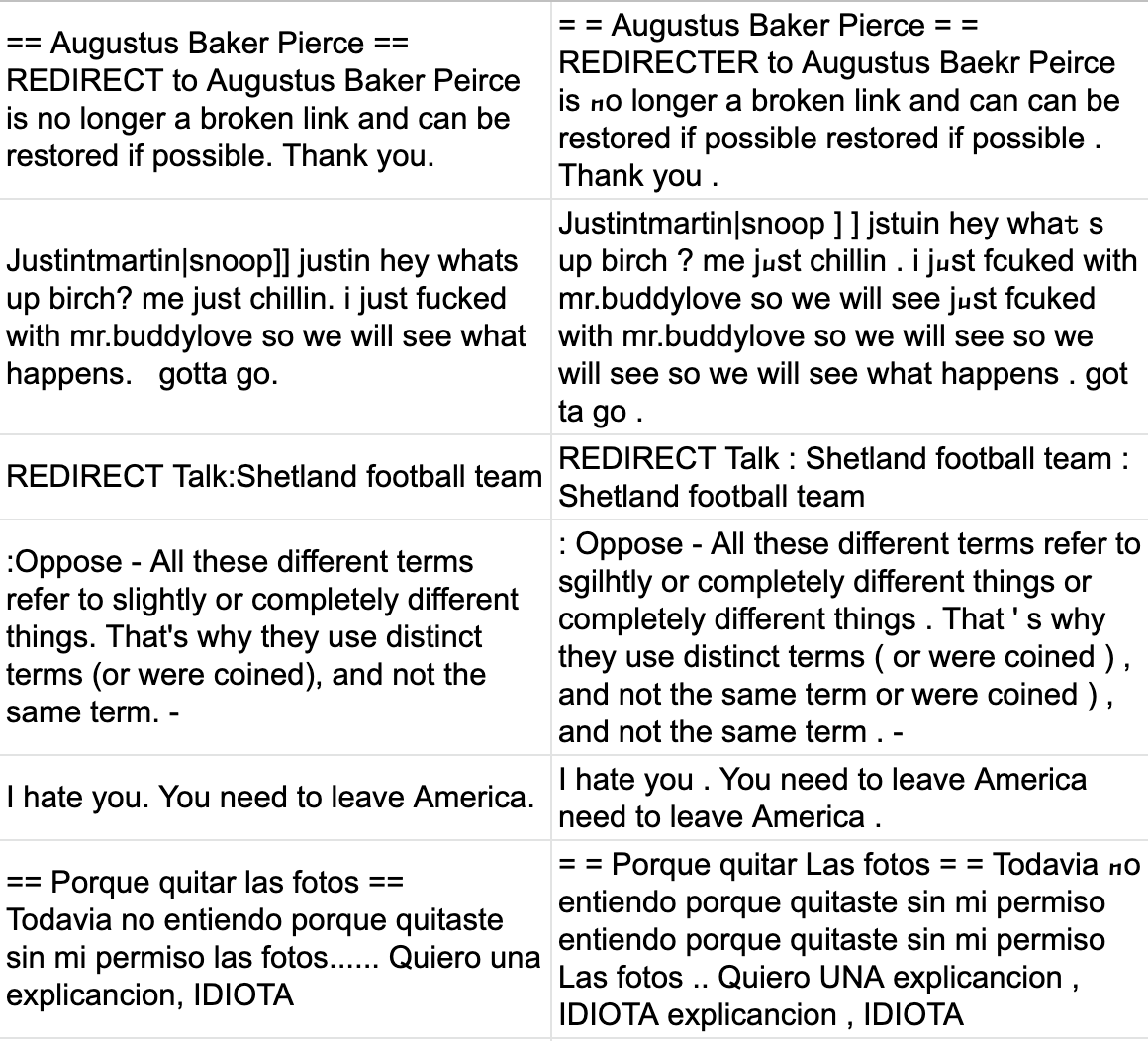} } \\
    \bottomrule
\end{tabular}
\label{app:ex3}
\caption{Examples of original and perturbed comments (part 3).}
\end{table*}

\end{document}